\documentclass[conference]{IEEEtran}
\IEEEoverridecommandlockouts
\usepackage{cite}
\usepackage{amsmath,amssymb,amsfonts}
\usepackage{array}
\usepackage[colorlinks=false, pdfborder={0 0 0}]{hyperref}
\usepackage{algorithmic}
\usepackage{graphicx}
\usepackage{textcomp}
\usepackage{xcolor}
\usepackage{listings}
\usepackage{caption}
\usepackage{subcaption}
\usepackage{setspace}
\usepackage{tabularx}
\usepackage{titlesec}
\usepackage{dashrule}
\usepackage{datetime}
\usepackage{times}
\usepackage{ragged2e}
\usepackage{tocloft}
\usepackage{chngcntr}
\usepackage{afterpage}
\usepackage{fancyhdr}
\usepackage{algorithmic}
\usepackage{textcomp}
\usepackage{makecell}
\usepackage{comment}
\usepackage{balance}
\usepackage{url}
\usepackage{orcidlink}
\def\BibTeX{{\rm B\kern-.05em{\sc i\kern-.025em b}\kern-.08em
    T\kern-.1667em\lower.7ex\hbox{E}\kern-.125emX}}
\begin{document}

\title{Towards Edge Artificial Intelligence for Predicting Antimicrobial Resistance in Mycobacterium tuberculosis 
\\
}
\author{
Nishat Tasnim\IEEEauthorrefmark{1},
Faisal  Ahmed\orcidlink{0000-0002-7601-3962}\IEEEauthorrefmark{2}\IEEEauthorrefmark{3}\thanks{Corresponding author: Faisal Ahmed}
,
Rokan Uddin Faruqui\IEEEauthorrefmark{1}
\\[0.5em]
\IEEEauthorblockA{\IEEEauthorrefmark{1} Department of Computer Science and Engineering, University of Chittagong, Bangladesh\\
Emails: tasnim.cse@std.cu.ac.bd, rokan@cu.ac.bd}
\IEEEauthorblockA{\IEEEauthorrefmark{2} Department of Computer Engineering and Mathematics, Universitat Rovira i Virgili, Spain}
\IEEEauthorblockA{\IEEEauthorrefmark{3} NVISION Systems and Technologies SL (INTERA Group), Barcelona, Spain\\
Email: faisal.ahmed@intera-group.com}
}
\maketitle

\begin{abstract}
Antimicrobial resistance (AMR) in Mycobacterium tuberculosis (MTB) poses a major challenge to global tuberculosis control, as multidrug-resistant strains limit available treatment options. Rapid and accurate prediction of drug resistance from genomic data could support timely treatment decisions and public health interventions. However, many existing computational approaches for genomic AMR prediction rely on computationally intensive pipelines, limiting their suitability for deployment on resource-constrained edge devices. This limitation is particularly relevant in hospital settings in low- and middle-income countries, where computing infrastructure and access to high-performance computational resources may be limited. To address this challenge, we present a lightweight, novel machine learning (ML) approach suitable for running on edge devices that combines \(4\)-mer frequency features derived from whole-genome sequences with a Random Forest classifier to predict antibiotic resistance in MTB. The proposed ML approach achieved an average area under the receiver operating characteristic curve of 0.82 across 20 antibiotics on an independent test set. 
\end{abstract}

\section{Introduction}
Tuberculosis (TB) remains one of the leading causes of death from infectious disease worldwide. The World Health Organization estimates that millions of new cases and over a million deaths occur each year \cite{who2024}. The bacterium \emph{Mycobacterium tuberculosis (MTB)} causes the disease, which is transmitted through airborne droplets released when infected individuals cough or sneeze. Drug-susceptible TB has a treatment success rate of approximately 88\%, but this drops to 57--68\% for multidrug-resistant and rifampicin-resistant cases. Global eradication efforts are therefore particularly hindered by the emergence and spread of drug-resistant strains, and rapid, accurate identification of antimicrobial resistance (AMR) is critical for patient care, treatment stewardship, and public health surveillance \cite{reller2009antimicrobial}.

Traditional antibiotic susceptibility testing relies on culture-based methods that require four to six weeks to produce results, delaying appropriate treatment and contributing to poor patient outcomes \cite{salam2023conventional}. Molecular diagnostics such as GeneXpert have been widely adopted for their speed and simplicity, but these assays detect only specific resistance-conferring mutations in targeted genomic regions and can therefore miss novel or rare variants, leading to misdiagnosis and ineffective treatment \cite{haas2025xpert}. These problems motivate the development of computational tools that capture the full picture of drug resistance and provide fast and accurate predictions. Machine learning (ML) has emerged as a powerful computational approach for predicting AMR from genomic data \cite{10.1093/bioinformatics/btab681,groschel2021gentb,paredes2025predicting}. 

ML-based approaches for predicting antibiotic resistance in \emph{Mycobacterium tuberculosis} typically rely on resistance-associated variants, such as single-nucleotide polymorphisms (TBProfiler \cite{phelan2019integrating}; Mykrobe \cite{hunt2019antibiotic}), or employ high-dimensional feature representations (MTB++ \cite{serajian2024scalable}). Methods based on resistance-associated variants require prior knowledge of resistance-associated mutations, limiting their ability to identify novel mutations \cite{lopezfernandez2024valencia,pal2025machine}. High-dimensional representations avoid this limitation but are computationally intensive, making them less suitable for deployment in resource-limited settings. This limitation is particularly relevant in high-burden, low-resource settings, where sequencing capacity is increasingly available, but sustained access to high-performance computing or reliable network connectivity to remote servers is not. In such settings, on-device inference could enable resistance prediction to remain local to the point of care. Additionally, most existing studies evaluate model performance on a restricted set of first-line antibiotics and often lack rigorous experimental designs.

To address these limitations, we present a lightweight ML approach for predicting AMR in MTB from whole-genome sequences using a compact \(4\)-mer frequency representation, designed for edge deployment in resource-constrained hospital and clinic settings. Our approach differs from previous research in two principal ways. First, it replaces the high-dimensional feature representations  with a 256-dimensional \(4\)-mer frequency vector, reducing the feature space by several orders of magnitude and removing the need for sparse matrix representations or large-scale feature ranking. Second, it evaluates resistance against 20 antibiotics, including several newer and repurposed drugs that have received limited attention in prior studies. 
Concretely, we extract 4-mer frequency vectors directly from whole-genome sequences, normalize them by total k-mer count, and use them as input to antibiotic-specific binary Random Forest (RF) classifiers. We train one classifier per antibiotic, with class weighting to compensate for the imbalance between resistant and susceptible isolates that affects many drugs. Our main contributions are as follows:
\begin{enumerate}
    \item We propose a lightweight approach for AMR prediction in MTB that combines a compact \(4\)-mer frequency representation with a random forest classifier targetting 
    a computationally simpler framework that is better suited to resource-constrained edge computing environments.
    \item We systematically evaluate the performance of the proposed method across a diverse range of antibiotics, including first-line, second-line, and new and repurposed drugs, using rigorous five-fold cross-validation and independent test sets. Our results demonstrate that \(4\)-mer representations can achieve robust predictive performance for well-characterized antibiotics.
\end{enumerate}

\section{Methodology}
\subsection{Feature encoding}

To construct the predictive model, raw DNA sequences were transformed into fixed-length numerical feature vectors using \textit{k}-mer feature encoding \cite{11490179} with $k = 4$. A 4-mer is defined as a contiguous nucleotide substring of length four. For each DNA sequence, a sliding window of length four with a step size of one nucleotide was applied to extract all overlapping 4-mers, yielding $(L - 3)$ 4-mers for a sequence of length $L$. Each extracted 4-mer was mapped to one of the $4^4 = 256$ possible nucleotide combinations formed from the four DNA bases (A, C, G, and T). The occurrences of all possible 4-mers were then counted to generate a 256-dimensional feature vector, where each feature corresponds to the frequency of a specific 4-mer.

To eliminate the influence of sequence length, the raw 4-mer counts were normalized by dividing each count by the total number of extracted 4-mers, thereby producing relative frequency values. Consequently, each DNA sequence was represented by a normalized 256-dimensional feature vector with values ranging between 0 and 1. This representation effectively captures the local compositional characteristics of genomic sequences while maintaining a fixed feature dimension across all samples.

Compared with smaller values of $k$, 4-mers capture richer sequence patterns and contextual information. Conversely, larger values of $k$ substantially increase the feature space, resulting in higher computational complexity and greater sparsity. Therefore, selecting $k = 4$ provides an effective balance between descriptive capability and computational efficiency, making it well suited for training a lightweight predictive model. Representative 4-mer frequency feature vectors generated from whole-genome sequences are presented in Table~\ref{tab:kmer_matrix}.

\begin{table}[htbp]
\centering
\caption{Representative 256-dimensional feature vectors of 4-mer frequency extracted from whole-genome sequences.}
\label{tab:kmer_matrix}
\resizebox{\columnwidth}{!}{%
\begin{tabular}{lcccccc}
\hline
Genome & 1 & 2 & 3 & $\cdots$ & 255 & 256 \\
\hline
1    & 0.042 & 0.038 & 0.041 & $\cdots$ & 0.042 & 0.038 \\
2    & 0.039 & 0.040 & 0.038 & $\cdots$ & 0.040 & 0.039 \\
$\vdots$    & $\vdots$ & $\vdots$ & $\vdots$ & $\ddots$ & $\vdots$ & $\vdots$ \\
598  & 0.039 & 0.041 & 0.040 & $\cdots$ & 0.040 & 0.039 \\
599  & 0.042 & 0.040 & 0.039 & $\cdots$ & 0.038 & 0.041 \\
600 & 0.038 & 0.041 & 0.042 & $\cdots$ & 0.041 & 0.040 \\
\hline
\end{tabular}}
\end{table}


\subsection{Predictive model construction}
To construct the predictive model, we used the Random Forest (RF) classifier because it is well established in genomic applications \cite{lim2021ienhancer,montesinos2022random}, more robust to collinear features than linear classifiers, and provides built-in feature importance estimates. It constructs multiple decision trees during training and aggregates their outputs by majority vote. This design limits overfitting, captures non-linear relationships between features and resistance phenotype, and yields importance scores that indicate which \(k\)-mer motifs contribute most to prediction .

We treated resistance prediction for each antibiotic as a separate binary classification problem and trained one classifier per antibiotic. The input was the \(256\)-dimensional normalized \(4\)-mer frequency vector, and the output was the corresponding binary phenotype (resistant or susceptible). Because the two classes were imbalanced for many antibiotics, we applied class weights to increase the cost of misclassifying the minority (resistant) class.

We selected hyperparameters through a preliminary grid search on the training set to balance predictive performance with computational cost. The final configuration used 100 estimators, a maximum depth of 10, and a minimum of 5 samples per split. 
\section{Experimental setup}
\subsection{Datasets}
The dataset used in this study was sourced from the Bacterial and Viral Bioinformatics Resource Center (BV-BRC)\footnote{\url{https://www.bv-brc.org/}}. AMR phenotype records for MTB were extracted, resulting in 329,980 resistant and 1,069,517 susceptible records. These records represent 755 genome sequences and 20 antibiotics. Each record comprises 21 attributes, such as genomic information, antimicrobial agents, and associated resistance phenotypes. For this study, we retrained only four attributes: Genome ID, Genome Sequence, Antibiotic, and Resistance Phenotype.

The records were subsequently grouped by antibiotic, resulting in 20 distinct subsets, each corresponding to a specific antibiotic. Each subset contains resistance phenotype labels for the respective antibiotic across the 755 genome sequences.

A comprehensive summary of the datasets is presented in Table \ref{tab:f1_correlation}.

\begin{table}[htbp]
\centering
\normalsize
\caption{Distribution of resistant (R) and susceptible (S) isolates per antibiotic}
\label{tab:f1_correlation}
\begin{tabular}{lrrc}
\hline
Antibiotic & R & S & R:S Ratio \\
\hline
Amikacin & 101 & 654 & 1:6.48 \\
Capreomycin & 137 & 618 & 1:4.51 \\
Ciprofloxacin & 128 & 592 & 1:4.63 \\
Clofazimine & 81 & 639 & 1:7.89 \\
Cycloserine & 82 & 638 & 1:7.78 \\
Delamanid & 448 & 272 & 1.65:1 \\
Ethambutol & 342 & 413 & 1:1.21 \\
Ethionamide & 143 & 577 & 1:4.03 \\
Isoniazid & 408 & 347 & 1.18:1 \\
Kanamycin & 165 & 590 & 1:3.58 \\
Levofloxacin & 285 & 435 & 1:1.53 \\
Moxifloxacin & 117 & 603 & 1:5.15 \\
Nicotinamide & 150 & 570 & 1:3.80 \\
Ofloxacin & 163 & 592 & 1:3.63 \\
Para-aminosalicylic acid & 62 & 583 & 1:9.40 \\
Prothionamide & 445 & 275 & 1.62:1 \\
Pyrazinamide & 331 & 389 & 1:1.18 \\
Rifabutin & 278 & 442 & 1:1.59 \\
Rifampin & 471 & 284 & 1.66:1 \\
Streptomycin & 400 & 355 & 1.13:1 \\
\hline
\end{tabular}
\end{table}

\subsection{Evaluation metrics}
We evaluated our method using the following standard classification metrics, where TP, TN, FP, and FN denote true positives, true negatives, false positives, and false negatives, respectively. Resistant isolates are treated as the positive class throughout.

\emph{Sensitivity (Sn)}, or recall, measures the proportion of resistant isolates correctly identified:
\begin{equation*}
Sn = \frac{TP}{TP + FN}
\end{equation*}

\emph{Specificity (Sp)} measures the proportion of susceptible isolates correctly identified:
\begin{equation*}
Sp = \frac{TN}{TN + FP}
\end{equation*}

\emph{Precision (Pr)} measures the proportion of predicted resistant isolates that are truly resistant:
\begin{equation*}
Pr = \frac{TP}{TP + FP}
\end{equation*}

\emph{F1-score (F1)} is the harmonic mean of precision and recall, combining both error types into a single value:
\begin{equation*}
F1 = 2 \times \frac{Pr \times Sn}{Pr + Sn}
\end{equation*}

\emph{Matthews correlation coefficient (MCC)} summarizes all four confusion-matrix entries and remains informative under class imbalance:
\begin{equation*}
MCC = \frac{(TP \times TN) - (FP \times FN)}{\sqrt{(TP + FP)(TP + FN)(TN + FP)(TN + FN)}}
\end{equation*}

\emph{Area under the ROC curve (AUC)} measures the model's ability to separate resistant from susceptible isolates across all classification thresholds.


\section{Results and Discussion}


\subsection{Cross-validation performance and robustness analysis}
\begin{table*}
\centering
\caption{Performance under 5-fold cross-validation, reported as mean $\pm$ SD (\%)}
\label{tab:cv_results}
\resizebox{\textwidth}{!}{%
\begin{tabular}{lcccccc}
\hline
Drug & Pr & Sn & Sp & F1 & MCC & 
AUC \\
\hline
Isoniazid     & 78.45 $\pm$ 2.72 & 75.16 $\pm$ 2.95 & 79.46 $\pm$ 3.78 & 76.70 $\pm$ 1.71 & 54.78 $\pm$ 3.43 & 85.34 $\pm$ 1.75 \\
Rifampin      & 88.22 $\pm$ 2.99 & 81.94 $\pm$ 3.74 & 84.45 $\pm$ 4.98 & 84.84 $\pm$ 1.06 & 65.97 $\pm$ 1.90 & 91.36 $\pm$ 2.19 \\
Cycloserine   & 80.88 $\pm$ 6.73 & 88.36 $\pm$ 7.12 & 97.95 $\pm$ 0.92 & 84.20 $\pm$ 5.24 & 82.87 $\pm$ 5.73 & 98.55 $\pm$ 0.60 \\
Prothionamide & 84.33 $\pm$ 3.94 & 83.15 $\pm$ 6.03 & 75.89 $\pm$ 7.44 & 83.55 $\pm$ 3.12 & 59.36 $\pm$ 7.11 & 89.60 $\pm$ 3.84 \\
Streptomycin  & 75.38 $\pm$ 2.90 & 69.77 $\pm$ 4.11 & 78.30 $\pm$ 3.99 & 72.34 $\pm$ 1.96 & 48.43 $\pm$ 3.56 & 84.04 $\pm$ 2.56 \\
Pyrazinamide  & 76.20 $\pm$ 2.52 & 67.37 $\pm$ 5.01 & 85.67 $\pm$ 0.91 & 71.48 $\pm$ 3.95 & 54.29 $\pm$ 5.48 & 85.60 $\pm$ 3.32 \\
Ethambutol    & 72.65 $\pm$ 3.91 & 69.91 $\pm$ 5.72 & 81.64 $\pm$ 3.31 & 71.14 $\pm$ 4.12 & 51.95 $\pm$ 6.11 & 86.07 $\pm$ 4.34 \\
Clofazimine   & 66.96 $\pm$ 9.00 & 61.67 $\pm$ 6.67 & 96.41 $\pm$ 1.38 & 63.89 $\pm$ 6.60 & 60.19 $\pm$ 7.33 & 93.12 $\pm$ 3.34 \\
Levofloxacin  & 73.35 $\pm$ 3.28 & 60.90 $\pm$ 5.12 & 87.49 $\pm$ 2.02 & 66.45 $\pm$ 3.86 & 50.76 $\pm$ 5.23 & 82.92 $\pm$ 2.97 \\
Kanamycin     & 76.06 $\pm$ 8.13 & 62.94 $\pm$ 5.05 & 95.53 $\pm$ 1.75 & 68.68 $\pm$ 5.21 & 63.10 $\pm$ 6.46 & 87.13 $\pm$ 3.08 \\
Rifabutin     & 76.09 $\pm$ 4.16 & 66.62 $\pm$ 6.79 & 88.89 $\pm$ 3.14 & 70.73 $\pm$ 3.70 & 57.63 $\pm$ 4.17 & 88.09 $\pm$ 0.94 \\
Ciprofloxacin & 49.47 $\pm$ 10.40 & 44.41 $\pm$ 12.57 & 92.50 $\pm$ 2.90 & 45.91 $\pm$ 9.94 & 38.52 $\pm$ 10.77 & 78.47 $\pm$ 3.83 \\
Delamanid     & 82.09 $\pm$ 4.60 & 77.81 $\pm$ 3.10 & 70.16 $\pm$ 8.25 & 79.87 $\pm$ 3.62 & 47.16 $\pm$ 10.19 & 81.59 $\pm$ 4.26 \\
Ethionamide   & 88.06 $\pm$ 4.52 & 65.33 $\pm$ 5.52 & 98.15 $\pm$ 0.77 & 74.93 $\pm$ 4.93 & 71.79 $\pm$ 5.42 & 81.53 $\pm$ 7.51 \\
Ofloxacin     & 52.46 $\pm$ 4.82 & 47.00 $\pm$ 5.10 & 91.20 $\pm$ 2.32 & 49.13 $\pm$ 1.31 & 39.96 $\pm$ 1.45 & 78.82 $\pm$ 4.31 \\
Capreomycin   & 61.12 $\pm$ 11.08 & 50.00 $\pm$ 12.17 & 94.12 $\pm$ 2.70 & 54.23 $\pm$ 10.74 & 47.90 $\pm$ 11.78 & 83.97 $\pm$ 5.08 \\
Moxifloxacin  & 48.67 $\pm$ 9.28 & 39.56 $\pm$ 6.53 & 94.23 $\pm$ 2.19 & 43.11 $\pm$ 6.25 & 37.02 $\pm$ 7.07 & 74.49 $\pm$ 7.81 \\
Nicotinamide  & 47.65 $\pm$ 5.82 & 37.11 $\pm$ 8.81 & 90.32 $\pm$ 2.04 & 41.31 $\pm$ 6.54 & 30.21 $\pm$ 6.71 & 73.77 $\pm$ 2.82 \\
Amikacin      & 67.41 $\pm$ 17.67 & 45.00 $\pm$ 8.50 & 97.04 $\pm$ 1.70 & 52.25 $\pm$ 6.10 & 50.03 $\pm$ 7.12 & 81.18 $\pm$ 6.50 \\
Para-aminosalicylic acid
              & 53.57 $\pm$ 11.07 & 29.72 $\pm$ 6.31 & 97.52 $\pm$ 1.05 & 37.81 $\pm$ 7.11 & 35.75 $\pm$ 7.84 & 76.08 $\pm$ 6.32 \\
\hline
Average & 69.95 $\pm$ 6.48 & 61.19 $\pm$ 6.35 & 88.85 $\pm$ 2.88 & 64.63 $\pm$ 4.85 & 52.38 $\pm$ 6.24 & 84.09 $\pm$ 3.87 \\
\hline
\end{tabular}%
}
\end{table*}

First, we evaluated model robustness using 5-fold stratified cross-validation and reported the mean and standard deviation of six evaluation metrics across folds (Table~\ref{tab:cv_results}). Model performance varied across antibiotics, indicating that predictive difficulty differed by drug. The highest performance was observed for rifampin, cycloserine, and prothionamide, each achieving F1-scores above 83\%. In contrast, prediction was less effective for para-aminosalicylic acid, nicotinamide, and several other drugs. This variation shows that performance differs across antibiotics and resistance phenotypes in the dataset.

Aggregate results show a consistent asymmetry in model performance: specificity was higher than sensitivity across the antibiotic panel. This indicates that the models were generally more effective at identifying susceptible isolates than resistant ones, even with class weighting. The imbalance was especially pronounced for antibiotics with fewer resistant isolates, where very high specificity coincided with comparatively poor recovery of resistance. Clinically, this distinction matters because false-negative resistance predictions may lead to inappropriate treatment, making sensitivity a particularly relevant metric for evaluating these models.

Variability in cross-validation results provides additional insight into where performance estimates should be interpreted with caution. Predictions for rifampin and isoniazid were relatively stable across folds, while greater fold-to-fold variation was observed for capreomycin, ciprofloxacin, and particularly amikacin. Amikacin, in particular, exhibited pronounced variability in precision despite a moderate mean value. This variability is consistent with the limited number of resistant isolates available for certain antibiotics. When the minority class is small, distributing resistant isolates across five folds can cause individual predictions to disproportionately influence fold-level metrics, resulting in large standard deviations. As a result, performance estimates for these antibiotics are less certain and may be more sensitive to the composition of the training and validation folds.

In summary, cross-validation results indicate that the model can provide robust resistance prediction for first-line antibiotics; however, performance is not uniform across the panel. The principal limitation remains the persistent gap between sensitivity and specificity, particularly for antibiotics with limited representation of resistant isolates. These findings underscore the importance of evaluating resistance prediction at the individual-antibiotic level and reporting class-sensitive metrics and their variability, rather than relying solely on aggregate performance. Furthermore, improving the representation of resistant isolates, especially for less prevalent resistance phenotypes, may be essential for enhancing both resistance detection and model generalizability.

\subsection{Performance analysis on the independent test set}

We further evaluated the model on an independent test set of 155 genomes not used during training. 

\begin{table}[h]
\centering
\caption{Performance on the independent test set (values in \%)}
\label{tab:test_results}
\resizebox{\columnwidth}{!}{%
\begin{tabular}{lcccccc}
\hline
Drug & Pr & Sn & 
Sp & F1 & MCC & 
AUROC \\
\hline
Isoniazid     & 89.19 & 90.00 & 73.33 & 89.59 & 63.76 & 89.36 \\
Rifampin      & 90.35 & 84.43 & 66.67 & 87.29 & 47.42 & 87.85 \\
Cycloserine   & 89.29 & 80.65 & 97.03 & 84.75 & 80.54 & 95.23 \\
Prothionamide & 83.91 & 82.02 & 67.44 & 82.95 & 48.91 & 84.11 \\
Streptomycin  & 87.50 & 77.06 & 73.91 & 81.95 & 47.96 & 84.30 \\
Pyrazinamide  & 90.28 & 70.65 & 82.50 & 79.27 & 49.06 & 84.58 \\
Ethambutol    & 80.00 & 75.00 & 69.49 & 77.42 & 43.78 & 82.57 \\
Clofazimine   & 64.29 & 85.71 & 90.99 & 73.47 & 68.63 & 93.52 \\
Levofloxacin  & 84.62 & 60.27 & 86.44 & 70.40 & 47.53 & 79.35 \\
Kanamycin     & 81.40 & 61.40 & 91.84 & 70.00 & 57.34 & 85.05 \\
Rifabutin     & 76.56 & 63.64 & 72.73 & 69.50 & 35.87 & 81.63 \\
Ciprofloxacin & 83.87 & 55.32 & 94.12 & 66.67 & 55.84 & 84.24 \\
Delamanid     & 67.61 & 64.86 & 60.34 & 66.21 & 25.10 & 69.79 \\
Ethionamide   & 91.30 & 50.00 & 97.78 & 64.62 & 58.67 & 79.09 \\
Ofloxacin     & 78.57 & 52.38 & 90.22 & 62.86 & 47.08 & 81.19 \\
Capreomycin   & 69.70 & 48.94 & 90.74 & 57.50 & 44.55 & 78.69 \\
Moxifloxacin  & 76.19 & 32.65 & 93.98 & 45.71 & 35.17 & 77.53 \\
Nicotinamide  & 46.43 & 35.14 & 84.21 & 40.00 & 21.25 & 71.44 \\
Amikacin      & 58.82 & 24.39 & 93.86 & 34.48 & 25.76 & 71.54 \\
Para-aminosalicylic acid
              & 42.86 & 16.67 & 95.96 & 24.00 & 19.21 & 79.04 \\
\hline

Average & 76.64 & 60.56 & 83.68 & 66.43 & 46.17 & 82.01 \\
\hline
\end{tabular}%
}
\end{table}
\begin{figure}
        \includegraphics[width= \columnwidth]{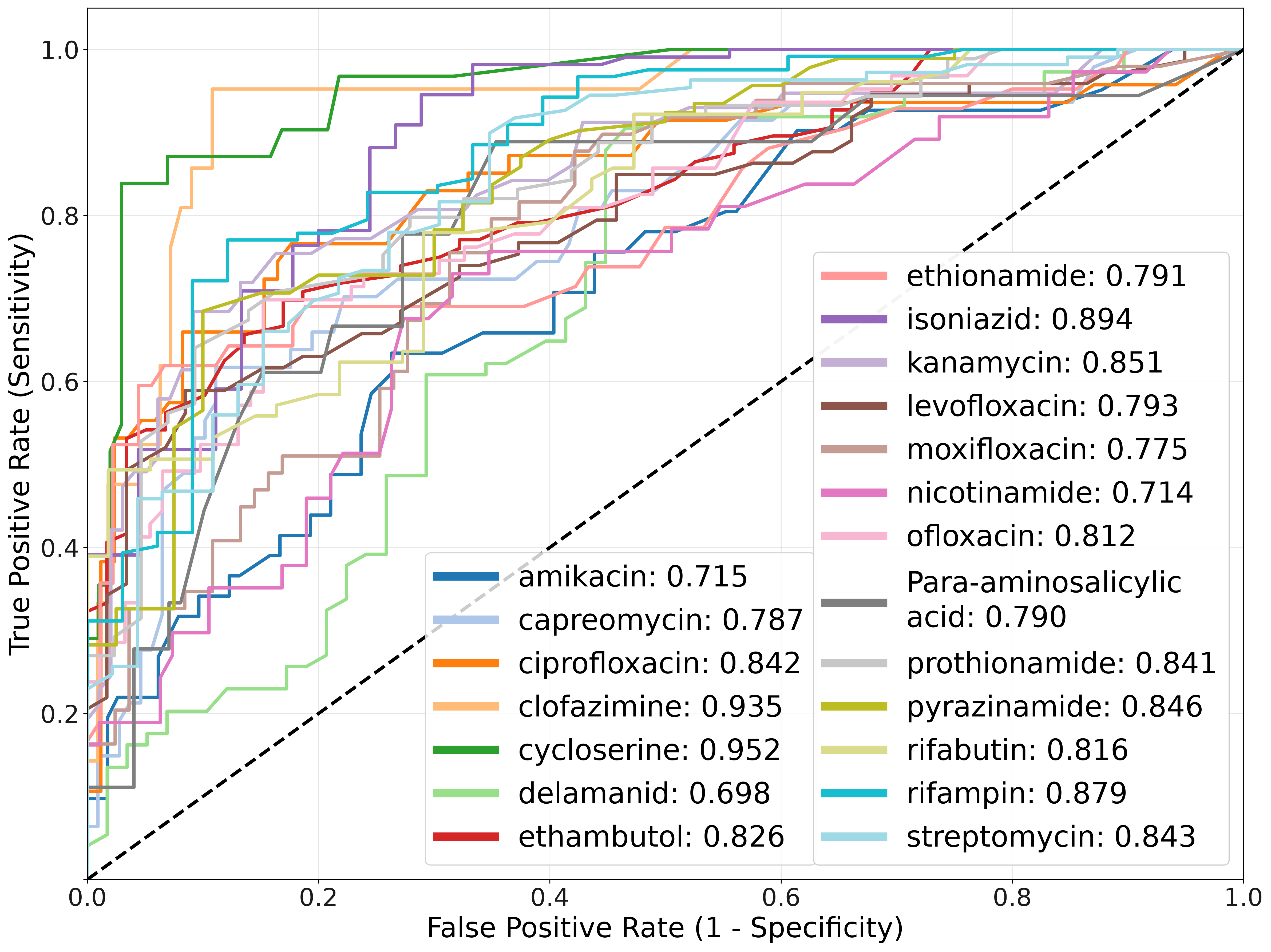}
        \caption{ROC curves for 20 antibiotics evaluated on the independent test set.}
        \label{fig:roc_all}
\end{figure}
Table~\ref{tab:test_results} reports the complete performance metrics for all 20 antibiotics on the independent test set. Across the 20 antibiotics, the model achieved an average accuracy of 77.50\%, an F1-score of 0.664, and a ROC-AUC of 0.820. Performance varied across antibiotics, with isoniazid and rifampin showing the strongest performance (F1-scores of 89.59\% and 87.29\%, respectively), while para-aminosalicylic acid and amikacin showed substantially lower performance.

Cycloserine remained among the best-performing antibiotics and showed strong overall discrimination, consistent with its cross-validation performance. These patterns suggest that predictive performance reflects the availability and distribution of resistance phenotypes.

Overall performance on the independent test set was lower than the cross-validation estimate, with accuracy decreasing from 84.30\% to 77.50\%. This difference is expected because cross-validation evaluates performance within the training-data distribution, whereas the independent test set provides a stricter assessment on previously unseen genomes. 

To visually complement the quantitative results presented in Table~\ref{tab:test_results}, Figure~\ref{fig:roc_all} shows the ROC curves for all 20 antibiotics. Our method demonstrates excellent discriminative performance for cycloserine, clofazimine, and isoniazid, achieving AUC values greater than 0.89. In contrast, the ROC curves for delamanid, nicotinamide, and amikacin lie closer to the diagonal reference line, indicating reduced discriminative performance. This diminished performance is likely attributable to the severe class imbalance in the training data.

\subsection{Evaluating resource requirements on the edge device}
We assessed the feasibility of the proposed model for edge deployment by measuring inference latency, memory consumption, energy usage, and model storage requirements on representative commodity edge hardware: an Intel Core i5-1035G1 CPU (4 cores, 8 threads) with 12 GB of RAM running Ubuntu 24.04.5 LTS.

The model requires 16.20 MB of storage and achieves a mean inference latency of 71.57 ms on the independent test set. It utilizes 55.46 MB of RAM and consumes 4327.23 J of energy for inference across all test set samples, demonstrating suitability for deployment on resource-constrained edge devices for rapid point-of-care antimicrobial resistance detection.

\section{Conclusions}
We evaluated a lightweight ML approach for predicting antibiotic resistance in MTB from whole-genome sequences. RF classifiers trained on a compact \(256\)-dimensional \(4\)-mer representation achieved robust predictive performance for well-characterized first-line antibiotics, demonstrating that a substantially reduced feature space can retain predictive power. Performance degraded, however, for new and repurposed drugs, where resistant phenotypes are scarce and class imbalance is severe. This confirms that data availability, rather than feature representation, is the principal limitation for these agents.

In future work, we will 
address the data scarcity affecting newer agents by expanding the training set with additional public repositories such as CRyPTIC and by exploring resampling and cost-sensitive learning strategies for imbalanced phenotypes.

\balance
\bibliographystyle{IEEEtran}
\bibliography{references}

\end{document}